\documentclass[letterpaper, 10 pt, conference]{ieeeconf}  

\IEEEoverridecommandlockouts                              

\usepackage{graphics} 
\usepackage{epsfig} 
\usepackage{amsmath} 
\usepackage{amssymb}  
\usepackage{bm} 
\usepackage[acronym]{glossaries}

\usepackage[flushmargin,hang]{footmisc} 

\usepackage{flushend} 
\usepackage{comment}	
\usepackage[utf8]{inputenc}

\usepackage{siunitx} 

\usepackage{pgfplots}
\pgfplotsset{compat=newest}
\usetikzlibrary{plotmarks}
\usetikzlibrary{arrows.meta}
\usepgfplotslibrary{patchplots}
\usepackage{booktabs} 
\usepackage{grffile}
\usepackage{url}
\usepackage{hyperref}
\pgfplotsset{plot coordinates/math parser=false}
\newlength\fwidth
\newlength\fheight

\DeclareUnicodeCharacter{202F}{\,} 

\usepackage{cite}

\def\epsgaiji#1{\leavevmode\kern-0.025zw\raise-.37zh\hbox{%
  \epsfile{file=#1,width=1.05zw}}\kern-0.025zw}
\newcommand{\MARU}[1]{{\ooalign{\hfil#1\/\hfil\crcr\raise.167ex\hbox{\mathhexbox20D}}}}

\usepackage{array}

\newglossaryentry{sample}{
    name={Sample},
    description={An example term used for illustration}
}

\newacronym{isru}{ISRU}{In-Situ Resource Utilization}
\newacronym{trl}{TRL}{Technology Readiness Level}
\newacronym{dem}{DEM}{Discrete Element Method}
\newacronym{fem}{FEM}{Finite Element Method}
\newacronym{ai}{AI}{Artificial Intelligence}

\title{\LARGE \bf
Design and Development of a Modular Bucket Drum Excavator for Lunar ISRU} 

\author{Simon Giel$^{1}$, James Hurrell$^{2}$, Shreya Santra$^{2}$,\\ Ashutosh Mishra$^{2}$,  Kentaro Uno$^{2}$, and Kazuya Yoshida$^{2}$
\thanks{$^{*}$This work was supported by JST Moonshot R\&D Program, Grant Number JPMJMS223B.}
\thanks{$^{1}$ S. Giel is with the Faculty of Aerospace Engineering and Geodesy, University of Stuttgart, Stuttgart 70569, Germany, $^{2}$James Hurrell, $^{2}$ S. Santra, $^{2}$A. Mishra,  $^{2}$K. Uno, and $^{2}$ K. Yoshida are with the Graduate School of Engineering, Tohoku University, Sendai 980--8579, Japan. (E-mail: \tt{st191373@stud.uni-stuttgart.de})  }%
\thanks{\textit{The corresponding author is S. Giel. This work was performed while S. Giel was visiting Tohoku University.
}
    }%
}%

\begin{document}

\maketitle
\thispagestyle{empty}
\pagestyle{empty}


\begin{abstract}
\gls{isru} is one of the key technologies for enabling sustainable access to the Moon. The ability to excavate lunar regolith is the first step in making lunar resources accessible and usable.
This work presents the development of a bucket drum for the modular robotic system MoonBot, as part of the Japanese Moonshot program. A 3D-printed prototype made of PLA was manufactured to evaluate its efficiency through a series of sandbox tests.
The resulting tool weighs \SI{4.8}{kg} and has a volume of \SI{14.06}{L}. It is capable of continuous excavation at a rate of \SI{777.54}{kg/h} with a normalized energy consumption of \SI{0.022}{Wh/kg}. In batch operation, the excavation rate is \SI{172.02}{kg/h} with a normalized energy consumption of \SI{0.86}{Wh} per kilogram of excavated material.
The obtained results demonstrate the successful implementation of the concept. A key advantage of the developed tool is its compatibility with the modular MoonBot robotic platform, which enables flexible and efficient mission planning. Further improvements may include the integration of sensors and an autonomous control system to enhance the excavation process.

\end{abstract} 


\section{Introduction}\label{introduction}

Currently, major space agencies are undertaking preparations to establish lunar bases, either for crewed missions or as scientific research stations \cite{mueller_lunar_2022}. A crucial enabling technology for ensuring sustainable access to the lunar surface is \gls{isru}, which significantly reduces both launch mass and the cost of resupply missions from Earth by utilizing materials indigenous to the Moon \cite{mueller_lunar_2022}, \cite{azami_mohammad_comprehensive_2024}, \cite{just_gh_parametric_2020}, \cite{pengzhang_etal_overview_2023}.

Lunar regolith contains a variety of valuable resources, including oxygen, silicon, iron, titanium, as well as water in the form of ice \cite{pengzhang_etal_overview_2023}, \cite{varozza_ben_overview_2024}, \cite{crawford_iana_lunar_2015}, \cite{hegde_u_analysis_2012}. These resources have potential applications in life support systems, the production of rocket propellant, and the fabrication of metallic structures required for habitat construction \cite{azami_mohammad_comprehensive_2024}, \cite{pengzhang_etal_overview_2023}, \cite{varozza_ben_overview_2024}, \cite{crawford_iana_lunar_2015}. The \gls{isru} process consists of three main steps: excavation of the regolith, beneficiation of the minerals, and extraction of the desired elements and molecules \cite{hadler_kathryn_universal_2020}. Preceding these steps is resource identification, and following them are storage and utilization. The tool developed in this study addresses the first step of this process—regolith excavation.

\begin{figure}[t]
    \centering
    \includegraphics[width=\linewidth]{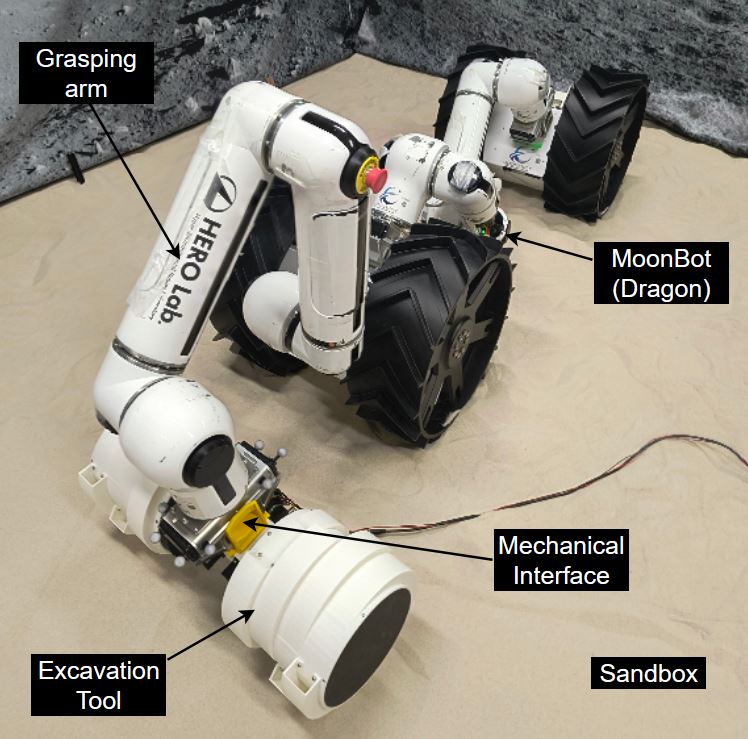}
    \caption{Assembly of the MoonBot in the Dragon configuration with the excavation tool attached to it.}
    \label{fig:dragon_tool_assembly}
\end{figure}

This work is part of a project under Goal 3 of the Japanese Moonshot Research and Development Program, which addresses societal challenges through innovation in robotics and artificial intelligence \cite{moonshot}. The project aims to create AI-driven robotic systems to support future manned lunar missions \cite{moonbot}. As part of this initiative, a modular robotic platform called MoonBot was developed, consisting of wheel and arm modules that can be assembled into various configurations. The configuration, comprising two wheels connected by an arm and an additional arm for manipulation, is referred to as the Dragon. It serves as the mobility platform for the excavation tool developed in this study. The Dragon equipped with the excavation tool is shown in Figure~\ref{fig:dragon_tool_assembly}.

The testing procedures, as well as the measured and calculated values, follow the recommendations outlined in \cite{just_gh_parametric_2020} to facilitate the comparison of different excavation concepts.

The main contributions of this work are the presentation of a design approach for a bucket drum excavator compatible with the MoonBot robots, the execution of various tests with the tool, the measurement of its performance, and the provision of comparable results.

\section{Related Works}\label{relatedwork}

The significance of \gls{isru} highlights the demand for efficient excavation technologies. Existing excavation systems can be broadly categorized into discrete and continuous excavation approaches, as well as into partial systems, comprising only the excavation tool, and integrated systems that include both the excavation mechanism and the mobility platform \cite{just_gh_parametric_2020}. Discrete systems, such as backhoes, scrapers, and front loaders, operate through individual cutting actions, after which the excavated material must be deposited. In contrast, continuous excavation systems, including bucket wheels, bucket drums, and bucket ladders, utilize multiple smaller cutting surfaces that sequentially engage the regolith.
Lunar excavation technologies must contend with challenges that differ significantly from terrestrial conditions due to the Moon's unique environment. One of the most critical differences is the reduced gravitational acceleration, which limits the amount of force a robotic excavator can exert on the surrounding terrain \cite{just_gh_parametric_2020}, \cite{mueller_rp_review_2012}, \cite{skonieczny_krzysztof_advantages_2016}. Continuous excavation methods are therefore particularly well suited for the lunar environment, as their small, sequential cutting actions reduce the reaction forces imposed on the rover.
Among continuous excavation approaches, the bucket ladder presents a particularly promising option, as it allows for direct regolith transport into a collection bin \cite{johnson_ll_excavation_2006}. This system comprises a series of buckets mounted on a continuous chain. While this design is effective under Earth-like conditions, the abrasive nature of lunar dust is expected to increase friction and degrade sliding components, thereby limiting operational lifespan \cite{stubbs_tj_impact_2004}. A novel approach introduced by \cite{skonieczny_krzysztof_advantages_2016} avoids the use of conveyor belts by employing a passive regolith transfer mechanism from a bucket wheel to a collection bin. The bucket drum represents another adaptation of the bucket wheel concept, consisting of a closed cylindrical drum with attached buckets \cite{just_gh_parametric_2020}. As it stores excavated regolith directly within the drum, it eliminates the need for additional material-handling subsystems, making it a particularly attractive option for lunar deployment.
Clark et al., in \cite{clark_dlarry_novel_2009} presents a bucket drum design that incorporates small internal baffles to prevent the regolith from spilling out once collected. The drum can be emptied by reversing its rotation direction. RASSOR 2.0 represents a robotic platform equipped with bucket drums mounted on both its front and rear ends \cite{mueller_robertp_design_2016}. The opposing configuration of the excavation units enables counteracting forces, effectively canceling out horizontal reaction forces during operation. The project is being further developed through the ISRU pilot excavator \cite{schuler_jason_m_isru_2022}. 
Furthermore, \cite{nakano_tomoyasu_dembased_2025} utilizes \gls{dem} simulations to refine the bucket drum design by identifying key performance-determining parameters. The buckets in this system are designed in a spiral configuration, forming a continuous 360° helical path that leads toward the drum's center. In this way, the buckets themselves constitute the structural body of the drum. \cite{li_haoran_multiobjective_2025} also presents an optimization of a bucket drum excavator design using \gls{dem}.



\section{Methodology}\label{methodology}

\subsection{System Overview}\label{system_overview}

In the present work, a design for a bucket drum excavator is proposed, based on its demonstrated potential as a highly efficient solution for lunar regolith excavation. The bucket drum system integrates excavation, material loading, and haulage into a single tool, thereby enhancing overall process efficiency. Its suitability for lunar operations is further supported by its continuous excavation mode, which minimizes reaction forces. Furthermore, the lack of secondary material transfer components gives an advantage in mitigating the abrasive and adhesive challenges posed by lunar dust.

\subsection{Hardware Configuration}\label{hardware}

The excavation tool consists of two drums connected via a central shaft. The mechanical interface to the MoonBot arm is located between the drums. To mitigate contamination of the bearings and gears by particles, felt seals were integrated at critical interfaces between moving components. 



\begin{figure}[b]
    \centering
    \includegraphics[width=0.9\linewidth]{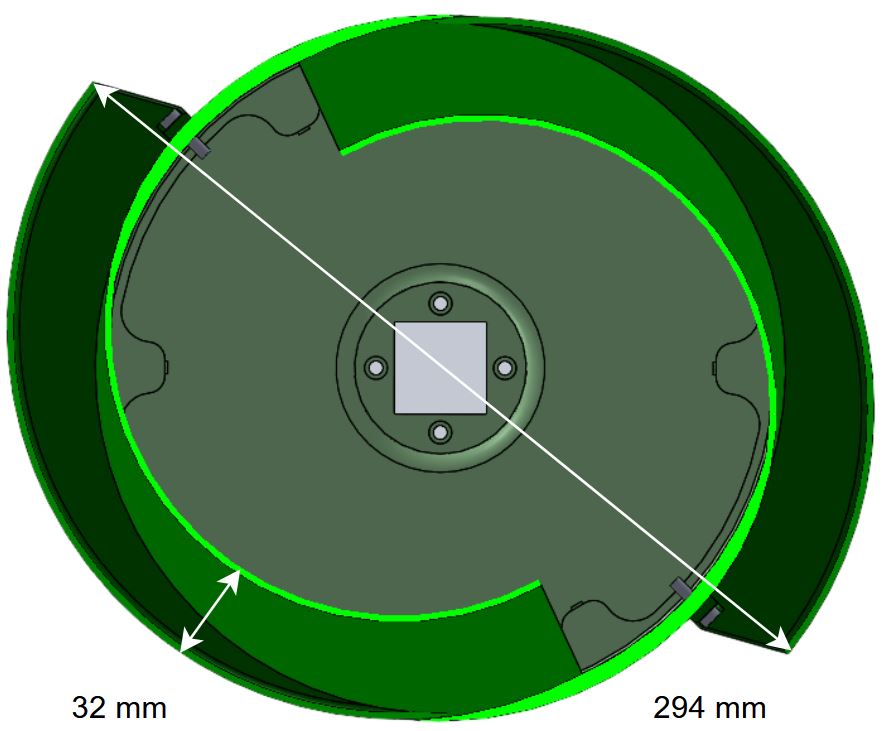}
    \caption{Section view of the CAD model of a drum featuring two buckets. Dark green indicates the buckets as separate parts, medium green shows the portions of the buckets that are integrated into the drum, and light green represents the cut surfaces revealed by the section view.}
    \label{fig:shovel design}
\end{figure}

The core component of the tool is the drum and bucket assembly, whose design plays a pivotal role in operational efficiency. A primary design challenge lies in preventing the loss of collected regolith, which can escape through the bucket inlets once stored inside the drum, reducing the effective storage capacity. In the present work, this issue is addressed by implementing spiral-shaped buckets that span an angular range of 160°. This configuration ensures that, during excavation, the bucket inlet remains at the drum’s lower edge while the outlet to the storage chamber stays above the level of the already collected material. This arrangement is intended to prevent material from falling out until the drum reaches a high fill level. The final drum and bucket configuration is illustrated in Figure~\ref{fig:shovel design}.


Each drum incorporates four pairs of opposing buckets. These pairs are rotationally offset to ensure that only one bucket is actively engaging with the regolith at any given moment, thereby minimizing instantaneous mechanical load and reducing reaction forces. The maximum penetration depth per cut is given by the height of the buckets, which is \SI{3.2}{cm}. To achieve a balanced distribution of forces during operation, the two drums are designed as mirror images of one another. The internal volume of the two drums, without the buckets, is \SI{14.06}{L}, increasing to \SI{20.96}{L} when the buckets are included.

All components of the excavation tool, except for the bearings, were manufactured using Fused Deposition Modeling with PLA filament. Components subjected to higher mechanical loads, such as the gears, were fabricated using PLA reinforced with carbon fibers to enhance strength and durability. The system is actuated by a \SI{24}{V}, \SI{50}{W} EC motor supplied by Maxon. The motor is coupled with a ceramic planetary gearhead featuring a reduction ratio of 936:1. Under maximum input speed, the resulting rotational velocity of the drums is approximately \SI{7.5}{rpm}.

\subsection{Electronics, Software and Control Framework} \label{software}

The motor is controlled via an EPOS4 Disk 60/8 motor driver. The motor’s power and Hall sensor lines are connected directly to the driver, which is powered by an external DC power supply. Communication between the motor driver and the control system is established via a USB connection. Motor operation was performed using Maxon’s EPOS Studio software, with the motor configured to run in Profile Velocity Mode for consistent speed control.

\subsection{Experiment Procedure}\label{experiment_procedure}

All tests were conducted in a \SI{3}{m} × \SI{4}{m} sandbox filled with Tohoku Silica Sand, designated as $"$Tohoku-Keisa No.~5" ~\cite{tohokukeisa}, to a depth of \SI{0.08}{m}. Preliminary manual experiments were performed before integrating the excavation tool with the Dragon. All experiments were carried out at the maximum motor speed.

The first test aimed to determine the maximum storage capacity of the bucket drum and the point at which material begins to fall back out of the drum through the buckets during excavation. To ensure the drum was fully loaded, the excavation process was continued for a period beyond the initial onset of material loss. Direct observation of the first instance of material falling out proved to be difficult, as the internal volume of the drum and the bucket configuration limit visibility. Therefore, an inverse method was employed: after filling the drum through excavation, the tool was lifted and rotated in the air without collecting additional material. As the rotation continued, the remaining sand gradually exited through the buckets. The point at which no further material was discharged was interpreted as corresponding to the initial point of material loss during excavation.

The subsequent test aimed to evaluate the excavation rate and power consumption under four different operational configurations. This test offers qualitative insights for comparative analysis of the tested parameters. The investigated configurations included: (1) standard excavation, (2) excavation with increased horizontal velocity, (3) excavation with increased applied force, and (4) reversed excavation, in which the excavation tool was moved such that the rotational direction of the buckets at the bottom of the drum opposed the tool’s horizontal motion. Each configuration was tested at least three times. The surface was flattened after each run to ensure consistent initial conditions.
The following parameters were recorded: excavation time, power consumption immediately before the end of the test run, the total weight of the excavation tool after material collection, and the horizontal distance traveled. In all configurations, the drum was only partially filled, up to a level below the previously determined threshold for material loss, to maintain a linear relationship between excavation volume and time.

\begin{figure}[t]
    \centering
    \includegraphics[width=\linewidth]{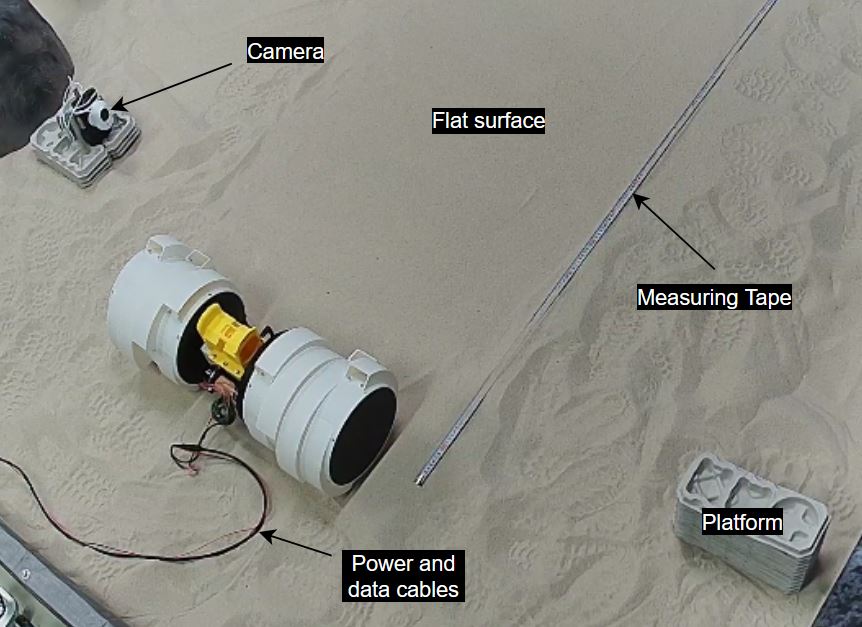}
    \caption{Test setup for the initial manual experiments, showing the leveled sand surface in the sandbox, the measuring tape, power and data cables, the unloading platform, and a side-mounted camera for visual monitoring and documentation.}
    \label{fig:test_setup_manual}
\end{figure}

A final manual test was carried out to determine the time and power requirements for unloading the drum. In this case, the prefilled drums from the previous experiment were placed on a stationary platform and unloaded. The measured variables were the time required to unload and the power consumption at the start of the unloading process. The experimental setup for the manual tests is shown in Figure~\ref{fig:test_setup_manual}.

For the tests conducted with the Dragon configuration of the MoonBot, the excavation tool was mounted to the rear of the rover. This configuration ensured that material was excavated behind the rover, preventing it from driving over the freshly disturbed soil. During testing, the tool was pressed into the ground by the rover with a moderate force. Once in contact, rotation of the tool was initiated, allowing it to sink a few centimeters into the soil. The rover then began moving forward while the excavation tool continued spinning. The total excavation time and the amount of excavated material were recorded during four test runs. The setup is illustrated in Figure~\ref{fig:dragon_tool_assembly}.

\section{Results and Analysis}\label{results}

The first manual test yielded a maximum fill mass of \SI{12.87}{kg} for the bucket drums. The onset of material loss, defined as the point at which regolith begins to fall back out of the drum, was observed at a fill mass of \SI{9.95}{kg}. The minimum bulk density of Tohoku Silica Sand was measured to be \SI{1.468e3}{kg/m^3}, following the methodology described in \cite{hurrell_traction_2025}. It should be noted, however, that this value was obtained using dry sand, whereas the sand in the test environment will have absorbed ambient moisture. As a result, the actual bulk density may have been slightly changed, leading to inaccuracies in the calculated volumes.

To determine the volumetric capacity of the system, it is necessary to define the adequate storage volume of the bucket drum. Although the buckets themselves are not primarily intended to retain material, a portion of the excavated regolith will naturally be stored within them during operation. To facilitate comparison with other excavation concepts, the volumetric capacity is presented for three volume configurations: Volume 1, the volume of the drums alone; Volume 2, the drums including the portion of the buckets located inside the drum; and Volume 3, the drums including the full volume of all buckets. For each of these configurations, the capacity is provided both at maximum loading and the point of first material loss in Table~\ref{tab:volumetric_capacity}. The onset of material loss is particularly relevant for evaluating the operational efficiency of the system, as the excavation rate can be considered approximately constant up to this point, but is expected to decrease significantly thereafter due to regolith loss.

\begin{table}[t]
\caption{Volumetric capacity at maximum loading and the onset of material loss for the three defined volume configurations.}
\label{tab:volumetric_capacity}
\centering
\begin{tabular}{@{}llll@{}}
\toprule
& \begin{tabular}[c]{@{}l@{}}Volume 1 \\ (\SI{14.06}{L}) \end{tabular} & \begin{tabular}[c]{@{}l@{}}Volume 2 \\ (\SI{18.39}{L}) \end{tabular} & \begin{tabular}[c]{@{}l@{}}Volume 3 \\ (\SI{20.96}{L}) \end{tabular} \\ \midrule
\begin{tabular}[c]{@{}l@{}}Volumetric capacity \\ at maximum loading [\%]\end{tabular}        & 62     & 48     & 42     \\
\begin{tabular}[c]{@{}l@{}}Volumetric capacity at \\ onset of material loss [\%]\end{tabular} & 48     & 37     & 32     \\ \bottomrule
\end{tabular}
\end{table}


The second manual test provided reference values for excavation rates under different operating configurations. For the standard excavation, the measured excavation rate was approximately \SI{777.54}{kg/h}, corresponding to a volume of \SI{0.53}{m^3/h}. It is important to note that these values are valid only up to the point at which material begins to fall out of the drums. Beyond this threshold, excavation becomes less efficient, leading to a reduction in the effective excavation rate. In the standard configuration, this threshold was reached after approximately \SI{46}{s}. Assuming a linear increase in power demand over time, the average power required for excavation was estimated at \SI{16.74}{W}. The normalized energy is then \SI{0.022}{Wh/kg}.

\begin{figure}[t]
    \centering
    \includegraphics[width=\linewidth]{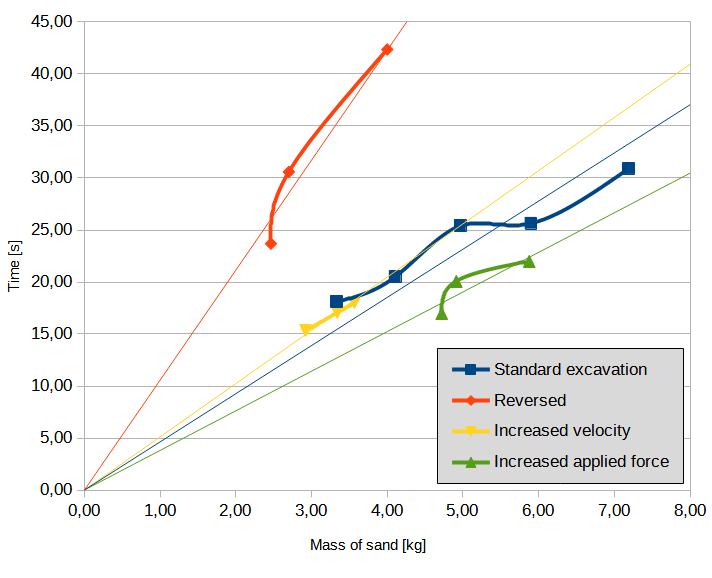}
    \caption{Comparison of the time required to excavate different amounts of Tohoku Silica Sand under various test conditions. Linear regression curves were fitted to the data and constrained to pass through the origin (0,0).}
    \label{fig:excavation_comparison}
\end{figure}

Figure~\ref{fig:excavation_comparison} presents a comparison of excavation performance across the tested configurations. The reversed excavation setup demonstrated significantly lower efficiency. Despite the potential advantage in reducing horizontal reaction forces on the rover, its low excavation performance makes it less attractive overall. The high-velocity configuration showed slightly reduced efficiency compared to the standard setup. At high speed the tool likely has limited ability to penetrate deeply into the regolith, resulting in only partially filled buckets. The high-force configuration exhibited the highest excavation rate, which can be attributed to improved penetration depth. Standard excavation was conducted at an average horizontal velocity of approximately \SI{2.7}{cm/s}, while high-velocity excavation was performed at \SI{7.2}{cm/s}. The applied force during the high-force test is estimated to be roughly twice that of the standard configuration. Although a force gauge was used for measurement, the results are not entirely accurate, as the effective force increases during filling due to the growing weight of the tool.


The final manual test provided reference values for the unloading process, specifically in terms of time and power requirements. The total time required to unload the drums, starting from the point at which material first begins to fall out, was \SI{23.3}{s}. The value for the unloading time was derived using a logarithmic regression function, which is considered appropriate, as fuller drums result in increased bucket fill levels during unloading, thereby accelerating the discharge process over time. The curve closely approximates an origin-passing logarithmic function, consistent with the expected unloading behavior in an ideal scenario. Assuming a linear trend in power consumption, the average power required for the unloading process was estimated to be \SI{14.13}{W}.




The final test provided data on the excavation rate of the tool when integrated with the Dragon. Table~\ref{tab:results_moonbot} summarizes the measured values and the calculated excavation rates. The average excavation rate was determined to be \SI{839.16}{kg/h}, which is comparable to the manually obtained value of \SI{777.5}{kg/h}. The slight increase is likely due to a higher applied contact force during the test runs.

\begin{table}[ht]
\caption{Measured values from the Dragon test and corresponding excavation rates. The degree of filling is defined relative to the previously defined Volume 2, which includes the drum and the parts of the buckets inside the drum.}
\label{tab:results_moonbot}
\centering
\begin{tabular}{@{}lllll@{}}
\toprule
Run                    & 1     & 2     & 3     & 4     \\ \midrule
Time [s]               & 22.80 & 39.14 & 23.52 & 28.99 \\
Loaded sand [kg]       & 4.96  & 8.98  & 5.64  & 7.08  \\
Degree of filling [\%] & 18    & 33    & 21    & 26    \\
Excavation rate [kg/s] & 0.22  & 0.23  & 0.24  & 0.24  \\ \bottomrule
\end{tabular}
\end{table}

Thus far, the presented results reflect only the performance of the partial system—the excavation tool. In order to evaluate the system's effectiveness within a complete \gls{isru} scenario, the mobility platform must also be considered. A complete operational cycle, referred to hereafter as batch operation, includes excavation, transport to a dumping area, unloading, and repositioning. For this analysis, the dumping area is assumed to be located \SI{10}{m} from the excavation site. The maximum travel speed of the Dragon is \SI{0.144}{m/s}. The estimated durations and power requirements for each step of the batch operation are summarized in Table~\ref{tab:batch_operation}.


\begin{table}[b]
\centering
\caption{Time and power consumption for each step of the batch operation cycle, including excavation, unloading, and driving.}
\label{tab:batch_operation}
\begin{tabular}{@{}llll@{}}
\toprule
          & Excavation & Unloading & Driving \\ \midrule
Time [s]  & 46.08      & 23.30     & 138.89  \\
Power [W] & 146.74     & 144.13    & 150    
\end{tabular}
\end{table}

The implementation of batch operation results in a reduced excavation rate of \SI{172.02}{kg/h}, corresponding to a volumetric rate of \SI{0.117}{m^3/h}. The normalized energy consumption amounts to \SI{0.86}{Wh/kg}. These values were calculated based on the results from the manual tests, as they closely match the excavation rates obtained with the Dragon. Additionally, no power consumption data was recorded during the tests with the Dragon, which further justifies the use of the manual test results for this analysis.

\section{Discussion}\label{discussions}

The results show that the developed concept achieves efficient regolith excavation, with an excavation rate of \SI{778}{kg/h} and a required energy of only \SI{0.022}{Wh} per kilogram of material for continuous operation. However, when comparing these results to the batch operation, where an excavation rate of \SI{172}{kg/h} and a normalized energy consumption of \SI{0.86}{Wh/kg} were observed, it becomes clear that the performance of the excavation tool alone does not determine overall system efficiency. The process chain as a whole must be optimized to improve throughput. One key aspect is minimizing the rover's travel time between the excavation and dumping sites, for example, by reducing the distance or optimizing the travel path.

It should also be noted that the presented results are based on linear filling up to the point of first material loss from the bucket drum. Excavation beyond this point leads to reduced efficiency in continuous operation due to slower filling rates. However, batch operation efficiency might improve in this case, as each cycle would transport more material while the increase in cycle time remains small.

Another critical factor influencing performance is the capability of the mobility platform. The current setup uses the Dragon, which, although highly modular and flexible, is not optimized for excavation tasks in terms of energy consumption, weight, or speed. Nonetheless, the modularity of the MoonBot platform opens up possibilities for optimizing system-level performance through alternative mission scenarios. For instance, a supporting cargo vehicle could periodically collect regolith from the excavating Dragon without requiring it to travel to a dumping area. This distributed excavation strategy, where multiple excavation units feed into a centralized transport unit, could significantly increase efficiency; however, it would also introduce complexity and reduce comparability with other concepts. Nevertheless, it represents a major advantage of the MoonBot project and highlights the benefits of its modular design.

Table~\ref{tab:comparison_BDE} provides a comparative overview of the proposed design and two existing bucket drum excavators. The results for this concept are largely comparable to those of RASSOR 2.0, particularly in terms of excavation rate and energy efficiency \cite{mueller_robertp_design_2016}, \cite{schuler_jm_rassor_2019}. However, direct comparisons remain challenging due to the absence of standardized test protocols. Variables such as traverse distance between excavation and dumping zones can greatly influence performance but are often defined differently in other studies \cite{just_gh_parametric_2020}. Furthermore, a standardized definition of bucket drum volume would be essential for making meaningful comparisons across different designs. The most suitable parameter for comparison would likely include both the volume of the drum itself and the combined volumes of all integrated buckets.

Additional parameters, such as reaction forces during operation, compaction metrics of the used simulant, and the volumetric capacity at the point of first material loss, should be considered in future studies. Differentiating between linear loading up to this point and total capacity would provide a clearer picture of operational efficiency and allow for more reliable benchmarking.

\begin{table}[t]
\caption{Comparison of the proposed bucket drum excavator with existing designs. Values marked with (c.) refer to continuous excavation; values marked with (b.) correspond to batch operation.}
\label{tab:comparison_BDE}
\centering
\begin{tabular}{@{}llll@{}}
\toprule
Parameters                                                             & This concept                                                        & RASSOR 2.0                                                        & \begin{tabular}[c]{@{}l@{}}Bucket drum\\ excavator\end{tabular}       \\ \midrule
\begin{tabular}[c]{@{}l@{}}Excavation\\ capability [kg/h]\end{tabular} & \begin{tabular}[c]{@{}l@{}}778 (c.)\\ 172 (b.)\end{tabular}         & 174 (b.)                                                          & \begin{tabular}[c]{@{}l@{}}1557-2016 (c.)\\ 389-504 (b.)\end{tabular} \\
\begin{tabular}[c]{@{}l@{}}Normalized\\ power [Wh/kg]\end{tabular}     & \begin{tabular}[c]{@{}l@{}}0.022 (c.)\\ 0.86 (b.)\end{tabular}      & 0.76 (b.)                                                         & 0.007-0.02 (c.)                                                       \\
\begin{tabular}[c]{@{}l@{}}System\\ mass [kg]\end{tabular}             & \begin{tabular}[c]{@{}l@{}}4.8 (tool)\\ 95 (total)\end{tabular}             & 67                                                                & 5.5 (tool)                                                            \\
\begin{tabular}[c]{@{}l@{}}Volumetric \\ capacity [\%]\end{tabular}     & 42-62                                                            & 50                                                             & 40                                                                \\
\begin{tabular}[c]{@{}l@{}}Traverse\\ speed [m/s]\end{tabular}         & 0.14                                                                & 0.49                                                              & -                                                                     \\
Simulant                                                               & \begin{tabular}[c]{@{}l@{}}Tohoku Silica \\ Sand \end{tabular} & \begin{tabular}[c]{@{}l@{}}BP-1 Regolith \\ simulant\end{tabular} & \begin{tabular}[c]{@{}l@{}}Regolith\\ simulant\end{tabular}           \\
Reference                                                              & -                                                                   & \cite{mueller_robertp_design_2016}, \cite{schuler_jm_rassor_2019} & \cite{clark_dlarry_novel_2009}                                        \\ \bottomrule
\end{tabular}
\end{table}

Although the prototype demonstrated proper functionality, several limitations were identified during testing. The absence of sensors—such as force and torque sensors, as well as devices to measure the amount of collected material—restricted the ability to perform more precise and quantitative evaluations. Additionally, an autonomous control system for regulating the depth of cut is required to enable integrated testing with the Dragon module.

While the current design considers certain aspects of the lunar environment, such as reduced gravity and, to some extent, lunar dust, it is not yet fully engineered to withstand the harsh conditions on the Moon, including extreme temperatures, abrasive regolith, and vacuum exposure. The presence of small stones and boulders that can get stuck in the inlet, as well as the compaction of the regolith in deeper layers, will negatively affect performance. All these aspects make an upgraded design necessary and could therefore increase the weight of the tool.

\section{Conclusions}\label{conclusion}

The bucket drum concept is inherently well suited for lunar operations due to its minimal number of mechanically moving parts and its ability to generate low reaction forces during excavation. These characteristics enhance its robustness in low-gravity environments and reduce the likelihood of failure caused by lunar dust contamination.

While the design of the bucket drum differs from other existing concepts, its overall performance is comparable. However, performance is determined not solely by the excavation tool itself, but mainly by the overall mission scenario, notably including the transport of excavated material to a designated dumping site. The integration of the modular MoonBot platform allows for alternative operational strategies, such as using dedicated transport rovers, thereby offering a distinct approach compared to existing concepts.

Future work should focus on optimizing the mechanical design through \gls{fem} simulations and refining the excavation parameters using \gls{dem} simulations. The implementation of force and material sensing, along with the development of an autonomous control system for regulating the depth of cut, is essential for conducting comprehensive and repeatable tests. Additionally, a detailed and realistic mission scenario should be defined and optimized to evaluate system performance under operational constraints.


\bibliographystyle{IEEEtran}
\bibliography{root.bib}

\end{document}